\documentclass[10pt,twocolumn,letterpaper]{article}

\usepackage[pagenumbers]{iccv} %
\usepackage[dvipsnames,table]{xcolor}
\usepackage{colortbl}  %
\usepackage{dsfont}
\usepackage{pifont}
\usepackage{multirow}
\usepackage{amsmath}
\usepackage{amsfonts}
\usepackage{amssymb}
\usepackage{algorithm}
\usepackage{algorithmic}
\usepackage{url}
\usepackage{booktabs}
\usepackage{subcaption}
\usepackage{float}
\usepackage{geometry}
\usepackage{tikz}
\usepackage{pgfplots}
\pgfplotsset{compat=1.16}

\newcommand{\para}[1]{\noindent\textbf{#1}}
\definecolor{paleBlue}{rgb}{0.8, 0.8, 1.0}
\definecolor{paleGreen}{rgb}{0.8, 1.0, 0.8}
\usepackage{graphicx}
\usepackage{cuted}

\usepackage{capt-of}
\usepackage{siunitx}

\definecolor{iccvblue}{rgb}{0.21,0.49,0.74}
\usepackage[pagebackref,breaklinks,colorlinks,allcolors=iccvblue]{hyperref}

\def\paperID{11844} %
\def\confName{ICCV}
\def\confYear{2025}

\title{Leveraging 2D Priors and  SDF Guidance for Dynamic Urban Scene Rendering}

\author{
Siddharth Tourani$^{2,3}$
\and
Jayaram Reddy$^{1}$
\and
Akash Kumbar$^{1}$
\and
Satyajit Tourani$^{1}$
\and
Nishant Goyal$^{5}$
\and
Madhava Krishna$^{1}$
\and 
N Dinesh Reddy$^{4}$
\and
Muhammad Haris Khan$^{2}$ \qquad\qquad
\and
\small{\textsuperscript{1} IIIT Hyderabad},
\small{
\textsuperscript{2}MBZUAI}, 
\textsuperscript{3}University of Heidelberg,
\textsuperscript{4} VLM Run, \textsuperscript{5} IIT Kharagpur\\
\small{\url{https://dynamic-ugsdf.github.io/}}\\
}
\begin{document}
\maketitle
\begin{abstract}
Dynamic scene rendering and reconstruction play a crucial role in computer vision and augmented reality. Recent methods based on 3D Gaussian Splatting (3DGS), have enabled accurate modeling of dynamic urban scenes, but for urban scenes they require both camera and LiDAR data, ground-truth 3D segmentations and motion data in the form of tracklets or pre-defined object templates such as SMPL. In this work, we explore whether a combination of 2D object agnostic priors in the form of depth and point tracking coupled with a signed distance function (SDF) representation for dynamic objects can be used to relax some of these requirements. We present a novel approach that integrates Signed Distance Functions (SDFs) with 3D Gaussian Splatting (3DGS) to create a more robust object representation by harnessing the strengths of both methods. Our unified optimization framework enhances the geometric accuracy of 3D Gaussian splatting and improves deformation modeling within the SDF, resulting in a more adaptable and precise representation. We demonstrate that our method achieves state-of-the-art performance in rendering metrics even without LiDAR data on urban scenes. When incorporating LiDAR, our approach improved further in reconstructing and generating novel views across diverse object categories, without ground-truth 3D motion annotation. Additionally, our method enables various scene editing tasks, including scene decomposition, and scene composition.
\end{abstract}

\begin{figure}
\centering
\includegraphics[width=0.5\textwidth]{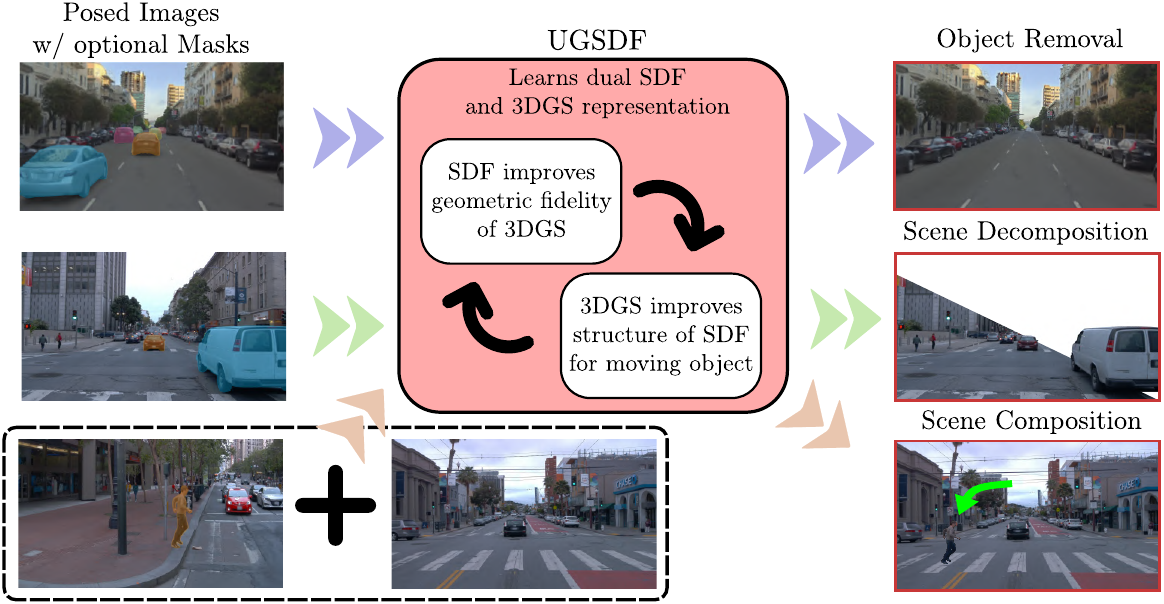}
\caption{We propose \textbf{U}rban \textbf{G}aussians via \textbf{S}igned \textbf{D}istance \textbf{F}unctions (UGSDF) for dynamic object modeling and rendering in urban scenes. UGSDF maintains a Signed Distance Function (SDF) and 3D Gaussian Splatting (3DGS) representation to model and render a dynamic object. It can be used for object removal, scene decomposition, scene composition (\textit{inserted object indicated with green arrow}) and other tasks related to simulation in urban scenes.}
\label{fig:teaser}
\end{figure}
\section{Introduction}
\label{sec:intro}

Representing and modeling large-scale dynamic scenes serves as the foundation of 3D scene understanding, playing a critical role in various autonomous driving tasks such as 3D object detection~\cite{cai2023objectfusion, chen2023futr3d}, motion planning~\cite{teng2023motion} and simulation of safety-critical scenarios~\cite{wang2021advsim,dosovitskiy2017carla}. As long as driving simulators like CARLA~\cite{dosovitskiy2017carla} and MARS~\cite{wu2023mars} work with synthetic data, there remains a big domain gap to real-world driving scenarios, which can lead to adverse results when deployed in self-driving cars~\cite{cruise2024}.

To mitigate the domain gap, real-world data integration is essential. Neural Radiance Fields (NeRFs)~\cite{mildenhall2021nerf} and 3D Gaussian Splatting (3DGS)~\cite{kerbl20233d} excel in scene reconstruction and novel view synthesis with high visual fidelity but may compromise geometric accuracy. In contrast, Signed Distance Fields (SDFs) offer precise surface modeling but require dense representations to achieve comparable visual fidelity.  

Separate from real-world verisimilitude, reconstructing dynamic driving scenes is challenging due to diverse actors, sensor noise, complex motions, occlusions, and so on.
Early methods to tackle these challenges focused on static scenes~\cite{tancik2022block,martin2021nerf,rematas2022urban,guo2023streetsurf}. Recent methods handle dynamic scenes by either \textbf{(i)} decomposing the scene into static and dynamic components via NeRFs~\cite{yang2023emernerf,turki2023suds,tancik2022block} and 3DGS~\cite{huang2024s3gaussians,zhou2024drivinggaussian,chen2023periodic} or \textbf{(ii)} constructing a scene graph~\cite{ost2021neural,kundu2022panoptic,yang2023unisim,wu2023mars,tonderski2024neurad,fischer2024multi,chen2024omnire} in which dynamic actors and static background are represented as nodes with edges encoding relative transformations that represent each actor's motion over time. \textit{Henceforth, we refer to any dynamic entity in a scene, including pedestrians as objects.}

Also of interest are methods for dynamic scene estimation ~\cite{som2024,lei2024mosca, wang2024gflow,liu2024modgs,yang2024deformable,luiten2024dynamic,xie2024physgaussian,teng2023motion, tourani2018mplp++, tourani2020taxonomy} that work on casually captured videos. These methods are object-agnostic and take as input \textit{point tracking} or optical flow data to capture object motion, depth, and camera poses for 3D information to render monocular sequences for novel viewpoints. Unlike these casually captured videos, traffic datasets typically have both camera and LiDAR data, as well as camera intrinsics, extrinsics, and object motion information in the form of object tracklets and for pedestrians SMPL templates. 

We introduce \textit{\textbf{U}rban \textbf{G}aussians via \textbf{S}igned \textbf{D}istance \textbf{F}unctions} (UGSDF) (~\cref{fig:teaser}),  a novel approach that reduces reliance on 3D object-specific motion annotations and LiDAR by leveraging 2D priors from  depth networks and point trackers to extract 3D information and motion cues. In addition to standard driving dataset inputs—images, intrinsics, extrinsics, object masks and optionally LiDAR, our method also incorporates object tracks from a point tracker and depth from a depth network. This enables state-of-the-art scene reconstruction and view synthesis without requiring object tracklets, 3D bounding boxes, or SMPL templates. UGSDF models dynamic objects using a combination of 3D Gaussian primitives and a Signed Distance Function (SDF), integrating their complementary strengths for enhanced rendering and geometric accuracy. The 3D Gaussians facilitate motion modeling and high-fidelity rendering, generating depth maps that refine SDF ray sampling. In turn, the SDF smooths surfaces and aids in Gaussian placement, iteratively improving the scene’s geometry. This dual representation effectively captures dynamic objects while remaining adaptable to static scenes by simply omitting motion modeling when unnecessary.

\noindent Our \textbf{contributions} are:
\begin{itemize}
    \item \textbf{2D Prior Based Dynamic Object Modeling In Urban Scenes} Our method derives motion and 3D structure information of dynamic objects from off-the-shelf point-trackers~\cite{karaev2023cotracker} and metric depth networks~\cite{piccinelli2024unidepth} by learning a dual SDF and 3DGS representation. It does so without using 3D tracklet like  motion information.

    \item \textbf{SDF based improvement of Gaussian primitive distribution} We train an SDF deformation network to model the geometry of a dynamic object, which we in-turn query to improve the localization of Gaussians on the dynamic object. Likewise, we use the location of the Gaussians to focus the ray-sampling of the SDF network. 
    \item \textbf{State-of-the-art results on Reconstruction and Novel View Synthesis of Dynamic objects} 
    The learned dual representation excels in scene reconstruction and novel view synthesis, surpassing template-free methods on KITTI~\cite{geiger2013vision} and Waymo~\cite{sun2020scalability}, and even outperforming template-based methods in some cases. We further demonstrate its versatility on casually captured datasets.

\end{itemize}

\section{Related Works}

\noindent\textbf{Dynamic Modeling of Urban Scenes}  
Neural representations~\cite{mildenhall2021nerf,barron2022mip,barron2021mip,mueller2022instant,Fridovich-Keil_2022_CVPR,kerbl20233d} have emerged as a dominant force in novel view synthesis, and have since been extended to dynamic scenes. \textit{Deformation} based methods like~\cite{pumarola2021d, park2021nerfies,Tretschk_2021_ICCV, park2021hypernerf,Cai2022NDR, wu20234d, yang2024deformable, huang2023sc,Xian_2021_CVPR,li2021neural,li2022neural,luiten2024dynamic, NEURIPS2022_d2, zhu2024motiongs}, warp  time-varying observations to a canonical space via a deformation network or input image timestamps (or latent codes) into neural representations. These techniques are typically limited to small scenes, making them less effective for dynamic urban environments. 

\textit{Dynamic Decomposition} based methods like~\cite{turki2023suds,yang2023emernerf,huang2024s3gaussians} have demonstrated reconstruction abilities for dynamic driving scenes, but are limited in control due to their using a single dynamic field for all scene objects. 

\textit{Scene Graph}-based methods, such as~\cite{ost2021neural, yang2023unisim, wu2023mars, tonderski2024neurad, fischer2024multi, chen2024omnire, yan2024street}, model dynamic objects using separate neural representations within a scene graph. These approaches require ground-truth motion data, along with 3D bounding box tracklets and 2D masks. Additionally, OmniRe~\cite{chen2024omnire} incorporates SMPL templates for modeling pedestrians. Our method shows using 2D priors in the form of point trackers and depth networks when combined with our dual representation of SDF networks and 3D Gaussians can also yield high-fidelity novel-view synthesis without requiring any 3D annotations, even without LiDAR data.

\para{Neural Surface Reconstruction Meets 3DGS} 
Neural rendering has advanced neural surface reconstruction~\cite{Oechsle2021ICCV,yariv2021volume,NEURIPS2021_neus,li2023neuralangelo,neus2,yariv2023bakedsdf}, using neural networks to represent scene geometry through occupancy fields or SDF values. Recent methods~\cite{li2023neuralangelo,Reiser2024SIGGRAPH,adaptiveshells2023} leverage hashed feature grids~\cite{mueller2022instant} for enhanced representation power, achieving excellent results. Hybrid techniques combining surface and volume rendering~\cite{turki2024hybridnerf,adaptiveshells2023,Reiser2024SIGGRAPH} improve both speed and quality. Approaches like~\cite{guedon2024sugar,xiang2024gaussianroom,yu2024gsdf} align 3D Gaussians with surfaces, while~\cite{Huang2DGS2024} enhances ray-splat intersections. NeuSG~\cite{chen2023neusgneuralimplicitsurface} and 3DGSR~\cite{lyu20243dgsrimplicitsurfacereconstruction} use 3D Gaussian Splatting with SDF fields for static scenes. To our knowledge, we are the first to combine SDFs and 3DGS for dynamic urban scenes for modeling individual dynamic objects.

\begin{figure*}[t]
    \centering
\includegraphics[width=0.95\linewidth]{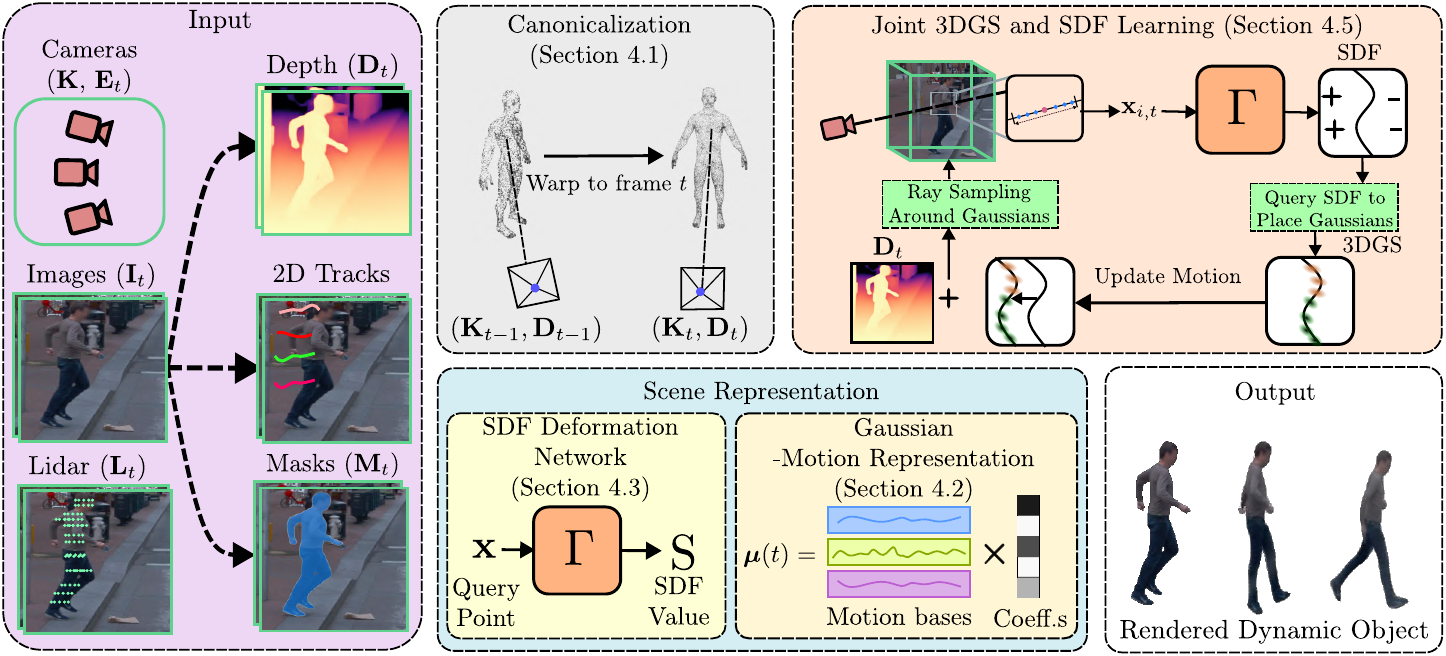}
    \caption{\textbf{Overview} UGSDF takes posed images, dynamic object masks, 2D tracking data, and depth maps (optionally LiDAR) as input, and outputs rendered dynamic scene. The initial model is constructed through canonicalization described in \Cref{subsec:init}. A dynamic 3DGS motion representation described in \Cref{subsec:dyn-repr}. The object of interest is represented using a combined representation of Signed Distance Functions (SDFs) and 3D Gaussian Splatting (3DGS) as detailed in \Cref{subsec:def-net} and the joint optimization scheme is described in~\Cref{subsec:sdf-3dgs}. These two representations are jointly learned. The coarse geometry of the Gaussians constrains the ray sampling of the SDF, while SDF queries add detail to the Gaussian representation.}
    \label{fig:overview}
    \vspace{-0.4cm}
\end{figure*}
 
\para{SAM Meets 3DGS} Semantic-NeRF~\cite{inplaceiccv2021} pioneered integrating semantic information into NeRF, enabling 3D segmentation from noisy 2D labels. Later work~\cite{fu2022panoptic,kundu2022panoptic,Siddiqui_2023_CVPR,wang2022dmnerf} introduced object-aware representations with instance modeling, relying on ground-truth labels. For open-world segmentation, approaches~\cite{tschernezki22neural,decomposing-nerf,Kerr_2023_ICCV,isrfgoel2023} distilled 2D features from foundation models~\cite{pmlr-v139-radford21a,li2022languagedriven,Caron_2021_ICCV} into radiance fields~\cite{zhou2024feature,qin2023langsplat}, but struggled with similar objects. SAM's open-world segmentation~\cite{Kim_2024_garfield,cen2023segment,ye2023gaussian,cen2023saga,kirillov2023segment,yang2024gaussianobject,ji2024segment,hu2024semantic} was adopted for applying 3D segmentation to static and dynamic but indoor scenes. In contrast, to the best of out knowledge, ours is the first method to use per-object SAM masks to learn 3D consistent segmentations in urban scenes without additional 3D annotations.

\section{Preliminaries} 
\label{sec:prelims}

\para{3D Gaussian Splatting} 3DGS~\cite{kerbl20233d} represents the 3D scene as differentiable 3D Gaussian primitives. Each Gaussian primitive is defined by the tuple $(\boldsymbol{\mu}, \mathbf{R}, \mathbf{s},\mathbf{o},\mathbf{r})$ where $\boldsymbol{\mu} \in \mathbb{R}^3$ is its mean, $\boldsymbol{R}$ and $\boldsymbol{S}$  are the orientation and scale of the Gaussians. $\mathbf{o} \in \mathbb{R}$ and $\mathbf{r} \in \mathbb{R}^3$ are the opacity and color (represented as the \textit{RGB} value) of the Gaussians. 
We denote the rendering operation $\Hat{\mathbf{I}} = \mathcal{R}(\mathcal{G}, \mathbf{K},\mathbf{E})$ where $\mathcal{R}$ is the rendering function,  $\mathbf{K}$, $\mathbf{E}$ are the camera intrinsics and extrinsics. $\mathcal{G}$ is the set of Gaussians and the output $\Hat{\mathbf{I}}$ is the rendered image. We can rasterize normal maps, depth maps and mask images by modifying the rendering equation as in~\cite{cheng2024gaussianpro,turkulainen2024dn}.

\para{Neural Implicit SDFs} Signed Distance Functions (SDFs) offer an effective way for representing surfaces implicitly as a zero-level set, $\{ \boldsymbol{x} \in \mathbb{R}^3 \mid \Gamma(\boldsymbol{x}) = 0 \}$, where $\Gamma(\boldsymbol{x})$ is the SDF value from a neural network $\Gamma(\cdot)$. Following NeuS \cite{NEURIPS2021_neus}, we replace the volume density with SDF and convert the SDF value to the opacity $\alpha_i$ with a logistic function:
\begin{equation}\small
\alpha_i = \max \left( \frac{\phi_s(\Gamma(\boldsymbol{x}_i)) - \phi_s(\Gamma(\boldsymbol{x}_{i+1}))}{\phi_s(\Gamma(\boldsymbol{x}_i))}, 0 \right),
\end{equation}
where $\phi_s$ denotes a Sigmoid function. Using the volume rendering methodology in~\cite{NEURIPS2021_neus}, the predicted color of pixel $\boldsymbol{p}$ can be used to render images. Likewise, replacing $\mathbf{c}_i$ with depth $\mathbf{d}_i$ and normals $\mathbf{n}_i$,

\section{Method}
\label{sec:method}
Our method \textit{Urban Gaussians via Signed Distance Functions} (UGSDF) takes as input a sequence of RGB frames captured by different cameras, represented as $\{\mathbf{I}_t^c \mid t=1, \dots, N\}$. Each frame includes a mask of the object of interest, $\{\mathbf{M}_t^c\}$, along with the camera intrinsics $\mathbf{K}^c \in \mathbb{R}^{3 \times 3}$ and world-to-camera extrinsics $\mathbf{E}^c_t \in \mathbb{SE}(3)$ for each frame $\mathbf{I}_t^c$ and camera $c$. When available, LiDAR scans $\mathbf{L}_t$ for each frame are also used as input. The goal of our method is to generate a coherent 4D representation and renderings of the object that maintain multi-view consistency, even in this sparse setup with limited viewpoints around the object. We assume the object of interest (denoted $\mathcal{O}$) appears in every frame, and we disregard any frames where $\mathcal{O}$ is absent. \Cref{fig:overview} gives an overview of our method. 

\noindent\textbf{Mask Generation} To generate dynamic masks $\{\mathbf{M}_t^c\}$, we combine point prompts in SAM2~\cite{ravi2024sam} with tracking from CoTracker~\cite{karaev2023cotracker}. The initial object mask is taken from the ground-truth annotations. A dense grid of points is tracked across the sequence using CoTracker, and the tracked points at each frame serve as prompts to SAM2, producing per-frame segmentations for each dynamic object. Finally, we post-process the points by pruning all point trajectories that lie outside the masks generated by SAM2.

\subsection{Construction of Canonical Model}
\label{subsec:init}
\para{Building the Initial Scaffold} We begin by constructing the initial object scaffold. We first compute the depth maps $\mathbf{D}_t^c$ of the images $\mathbf{I}_t^c$ using the pre-trained metric depth network UniDepth~\cite{piccinelli2024unidepth}. We use this network, as it handles the scale disparity for the depth maps between different frames. We then compute the object point clouds $\mathbf{P}_t$ for each of the different time-steps by lifting pixels corresponding to the $\mathcal{O}$ to 3D by backprojection, \ie if $\boldsymbol{p}_i \in \mathbb{R}^2$ is a pixel in $\mathbf{M}_t^c$, then $\mathbf{x}_i = \mathbf{D}_t^c(\boldsymbol{p}) \times  (\mathbf{K}^{c})^{-1}\tilde{\boldsymbol{p}}_i$, where $\tilde{\boldsymbol{p}}_i$ is $\boldsymbol{p}$ in homogeneous coordinates. We then warp the object point clouds to a single coordinate frame referred to as the canonical frame (typically the first frame where the whole object appears) by computing the 2D pixel tracks between the current frame and the canonical frame using~\cite{karaev2023cotracker}. This is done to warp point clouds across timesteps and also cameras within the same timestep for a multi-camera setting. We can likewise use the 2D tracks to warp the lidar point clouds corresponding to the $\mathcal{O}$ when available. We denote the object point cloud in the canonical frame obtained by this process as $\mathbf{P}_o$. Without loss of generality we drop the camera index $c$ for less cluttered notation.  Since this initial scaffold is based on noisy and partial observations of the object of interest, refinement is needed to prevent error propagation to subsequent frames. Thus, the scaffold initializes the Gaussian representation, which is then used to perform an initial training for the SDF network. Warping is done in a window of $5$ time-steps from the initial detection of the object to the inital detection time-step.

\para{Gaussian Initialization} We then initialize the 3D Gaussians of $\mathcal{O}$ in the canonical frame, where each 3D Gaussian $\boldsymbol{g}_o=(\mu_o, \mathbf{R}_o, \boldsymbol{s}, \boldsymbol{o}_o, \boldsymbol{r}_o)$ denotes the Gaussians, where $\mu_o \in \mathbb{R}^3$ are the means and are initialized with the points in $\mathbf{P}_o$. $\mathbf{R}_o$ are the orientation, $\boldsymbol{s}$ the scale, $\boldsymbol{o}_o$ the opacity, and $\boldsymbol{r}_o$ the color of the Gaussians. We denote $\mathcal{G}_o$ as the set of initialized Gaussians, \ie $\mathcal{G}_o = \{\boldsymbol{g}_o\}$. We denote the rendered image for extrinsic $\mathbf{E}_t$ and intrinsic $\mathbf{K}$ as $\Tilde{I}_o = \mathcal{R}(\mathcal{G}_o,\mathbf{E}_t, \mathbf{K}^c)$, where $\mathcal{R}$ is the differentiable rasterizer~\cite{kerbl3Dgaussians}. The parameters of $\mathcal{G}_o$ are then learned by minimizing
\begin{equation}
\small
    \mathcal{L}_{gs} = \mathcal{L}_{L1}(\mathbf{I}_o(\mathbf{M}_o), \mathcal{R}(\mathcal{G}_o))+ \mathcal{L}_{ssim}(\mathbf{I}_t(\mathbf{M}_t),\mathcal{R}(\mathcal{G}_o))
\end{equation}
where $\mathcal{R}_o(\mathcal{G}_o)$ denotes the rendered image and $\mathbf{I}_o(\mathbf{M}_o)$ is the part of the image corresponding to the $\mathcal{O}$.

\begin{figure}
    \centering
    \includegraphics[width=0.9\linewidth]{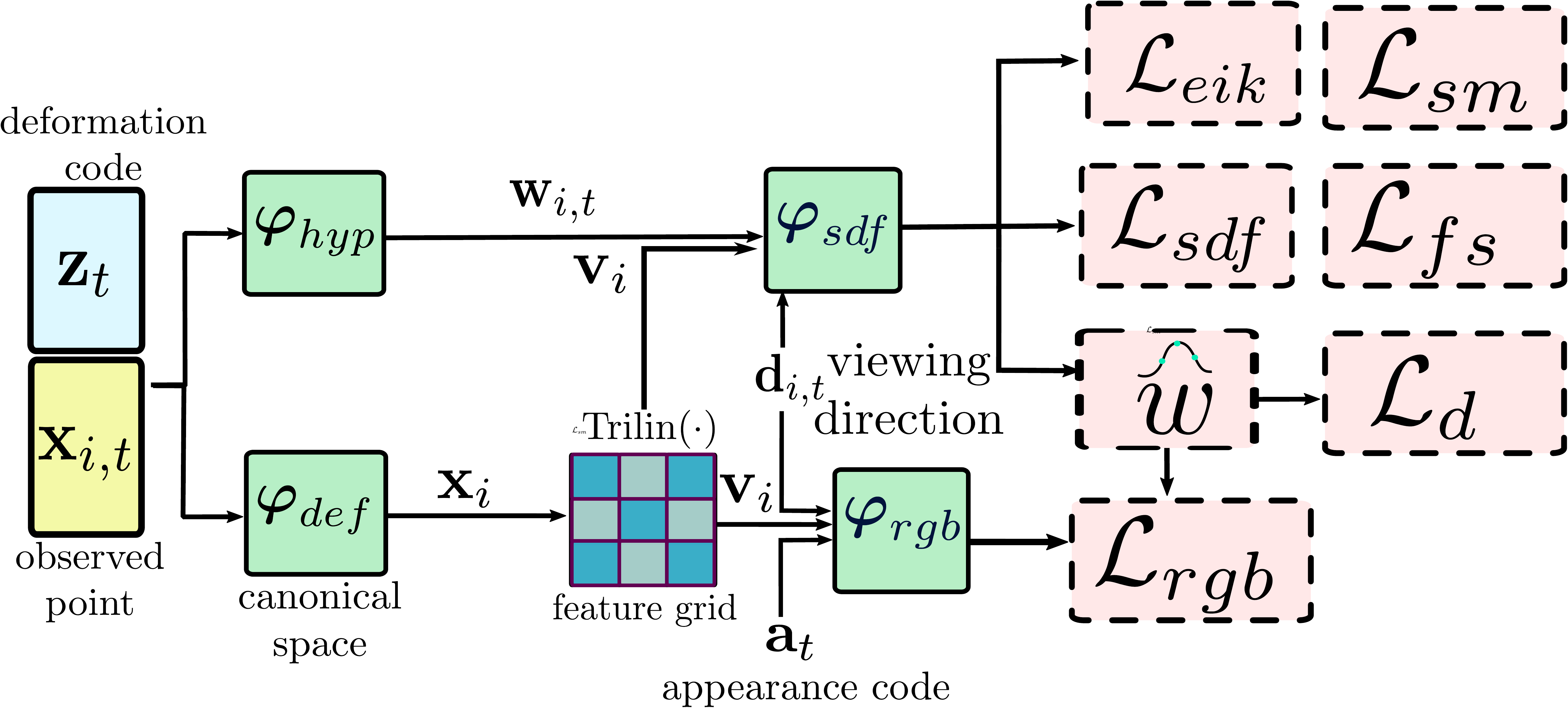}
    \caption{\textbf{SDF Deformation Network} The network takes as input the observed point $\mathbf{x}_{i,t}$ and outputs an SDF value $S_i(t)$ via the $\varphi_{sdf}$ MLP.}
    \label{fig:sdf-arch}
\end{figure}

\subsection{Gaussian Motion Representation}
\label{subsec:dyn-repr}

Since it is challenging to independently model the motion of each Gaussian $\boldsymbol{g}_o$, we adopt a common assumption (used in~\cite{yang2024deformable, som2024, abou2024particlenerf, li2023dynibar}) that models the motion trajectory as a linear combination of learnable basis trajectories. 
\begin{align}\small
    \boldsymbol{\mu}(t) = \boldsymbol{\mu}_o + \sum_{j=1}^B \mathbf{c}_j(t) \mathbf{b}^\mu_j(t) \label{eq:mean-basis}\\ \small \boldsymbol{q}(t) = \mathbf{q}_o + \sum_{j=1}^B \mathbf{c}_j(t) \mathbf{b}^R_j(t)\label{eq:rot-basis}
\end{align}
Here $B$ the number of basis trajectories is a hyper-parameter, $\mathbf{b}^\mu_j(t)$ and $\mathbf{b}^R_j(t)$ are learnable basis trajectories for the means $\mu_o \in \mathbb{R}^3$ and quaternions $\boldsymbol{q}_o \in \mathbb{R}^4$ respectively. The motion coefficients  $\{\mathbf{c}_j(t) \in \mathbb{R} \}_{j=1}^B$ are shared across means and rotations. Both quaternion $\mathbf{q}_o$ and $\mathbf{R}_o$ represent the same rotation  for each of the Gaussians. We additionally impose a sparsity penalty on the motion coefficients, forcing the Gaussian motion to be modeled by only a few basis trajectories to learn generalizable trajectories.

\subsection{SDF Deformation Network}
\label{subsec:def-net}
 We model the SDF function for $\mathcal{O}$ as a combination of a multi-resolution feature grid~\cite{mueller2022instant} and shallow MLPs. The architecture of the network is given in~\Cref{fig:sdf-arch} and is a modified version of the architecture used in~\cite{mohamed2024dynamicsurf}.
The feature grid consists of voxel grids of increasing resolution, each containing learnable features. To evaluate, each grid is queried via trilinear interpolation and the resulting features from all grids are concatenated.   
The network takes as input a learnable deformation code $\mathbf{z}_t$ and input point $\mathbf{x}_{i,t} \in \mathbb{R}^3$. These are both input into two different MLP networks: (1) \textit{the deformation network} $\boldsymbol{\varphi}_{def}$ that outputs the 3D point corresponding to $\mathbf{x}_{i,t}$ in canonical space. \ie $\mathbf{x}_{i} = \boldsymbol{\varphi}_{def}(\mathbf{x}_{i,t},\mathbf{z}_t)$. We drop the $t$ subscript for the canonical frame. (2) \textit{the topology-aware network} $\boldsymbol{\varphi}_{hyp}$ which outputs a higher dimensional mapping $\mathbf{w}_{i,t} \in \mathbb{R}^m$ where $m$ is the higher-dimensional space, \ie $\mathbf{w}_{i,t} = \boldsymbol{\varphi}_{hyp}(\mathbf{x}_{i,t},\mathbf{z}_t)$ These topology aware networks~\cite{park2021hypernerf} are designed to handle varying topologies by mapping the canonical shape to a higher dimension. $\mathbf{x}_{i}$ is then input into the multi-resolution feature grid $\mathcal{V}$ obtaining the tri-linearly interpolated features $V^l(\mathbf{x}_{i})$, at different resolutions ($l$ is over different resolutions). We denote the concatenation of the multi-resolution features $\{V^l(\mathbf{x}_{i}) \}$ as $\mathbf{v}_{i}$. The SDF output is obtained as $S_i = \boldsymbol{\varphi}_{sdf}(\mathbf{v}_{i},\mathbf{w}_{i,t})$ where $\boldsymbol{\varphi}_{sdf}$ is an MLP. Likewise color is obtained as $\mathbf{c}_i^t = \boldsymbol{\varphi}_{rgb}(\mathbf{v}_{i,o},\mathbf{w}_{i,t}, \mathbf{d}_{i,t},\mathbf{a}_t)$. The color MLP $\boldsymbol{\varphi}_{rgb}$ takes as input in addition to $\mathbf{v}_{i}$ and $\mathbf{w}_{i,t}$ the viewing direction $\mathbf{d}_{i,t} \in \mathbb{R}^3$ and appearance code $\mathbf{a}_t$. We abstract away the details of the network and just denote the network as $\Gamma$, which can be thought to take input $\textbf{x}_{i,t}$ and output a corresponding SDF Value $S_\mathbf{i}(t)$.

\subsection{Optimization}
\label{subsec:opt}
\para{SDF Loss} The SDF network is trained by minimizing:
\begin{equation}
    \mathcal{L}_{tot} = \mathcal{L}_{rgb} + \mathcal{L}_{d} +  \mathcal{L}_{sdf} + \mathcal{L}_{fs} + \mathcal{L}_{eik} + \mathcal{L}_{sm}
\end{equation}
where $\mathcal{L}_{rgb}$ and $\mathcal{L}_{d}$  are the  The RGB and depth per-pixel rendering losses, $\mathcal{L}_{fs}$ is the free-space loss~\cite{isdf2022,wang2022go-surf},  $\mathcal{L}_{eik}$ and $\mathcal{L}_{sm}$ are the eikonal regularization~\cite{pmlr-v119-gropp20a,isdf2022,NEURIPS2021_neus} and smoothness loss.  $\mathcal{L}_{eik}$ and $\mathcal{L}_s$ both regularize the surface to be smooth in the absence of point information. $\mathcal{L}_{fs}$ enforces the network to predict large SDF values between the camera origin and the observed surface. 

 \para{3DGS Loss} We supervise the dynamic Gaussians with two sets of losses. The first set of losses minimize discrepancy between the per-frame pixelwise
color, depth, and masks inputs. During each training step, we render the image $\Hat{\mathbf{I}}_t$, depth $\Hat{\mathbf{D}}_t$, and mask $\mathbf{M}_t$ from their training cameras $(\mathbf{K}, \mathbf{E}_t)$ via the differentiable rasterizer $\mathcal{R}$. These predication are supervised as:
\begin{align}
    \mathcal{L}_{L2}(\Hat{\mathbf{I}}_t,\mathbf{I}_t) + \mathcal{L}_{L2}(\Hat{\mathbf{D}}_t,\mathbf{D}_t) + 
    \mathcal{L}_{L2}(\Hat{\mathbf{M}}_t,\mathbf{M}_t) 
\end{align}
where $\mathbf{M}_t$, $\mathbf{D}_t$ and $\mathbf{I}_t$.  $\mathbf{I}_t$ is the original image, $\mathbf{D}_t$ is generated from UniDepth and $\mathbf{M}_t$ from SAM2~\cite{ravi2024sam}. The second set of losses impose the standard constraints on motion using 2D tracks and depth as well as rigidity taken from~\cite{som2024}.  The training is split into two stages, the initialization which trains the parameters of the Gaussians as well as the SDF network. We alternate between training the Gaussians and SDF network. We provide architecture and  implementation details in the \textbf{\textit{supp}}.

\subsection{Joint 3DGS and SDF Learning}
\label{subsec:sdf-3dgs}

 Inaccurate underlying geometry on dynamic objects cause Gaussians to be rendered as  floaters around the moving object. The original 3DGS~\cite{kerbl3Dgaussians} restricts its densification strategy solely to local operations like cloning or splitting, thereby posing challenges when generating new Gaussian primitives during densification in areas lacking  Gaussians already. Methods like~\cite{guedon2024sugar,cheng2024gaussianpro,xiang2024gaussianroom} steer the Gaussians to be close to planar surfaces, which work well for indoor scenes, as we are interested in handling non-planar moving objects we instead use the depth map to densify Gaussians. 

\para{SDF Guidance for Densification} Let $\mathcal{G}_t$ denote the set of Gaussians at time $t$ evolved via~\cref{eq:mean-basis} and~\cref{eq:rot-basis}. We render $\Hat{\mathbf{I}}_t = \mathcal{R}(\mathcal{G}_t, \mathbf{K},\mathbf{E}_t)$ and from it generate depth map $\Hat{\mathbf{D}}_t$ via UniDepth. We partition the region around the object into $N^3$ cubic grids. The dimensions of the region around the object are based on the span of the object model after canonicalization (denoted $\mathbf{S}_c$). Then, we query the SDF value at the center of each grid. SDF values   below the threshold $\tau_s$ indicates the grid is in proximity to the scene surface. In case, the Gaussians within the grid are below a certain number $N_g < \tau_n$ we back-project the depth map $\mathbf{D}_t$ creating a point cloud and  sample 3 points closest to $\mathbf{c}$ to densify the Gaussians.

\begin{figure}[t]
\centering\includegraphics[width=0.90\linewidth]{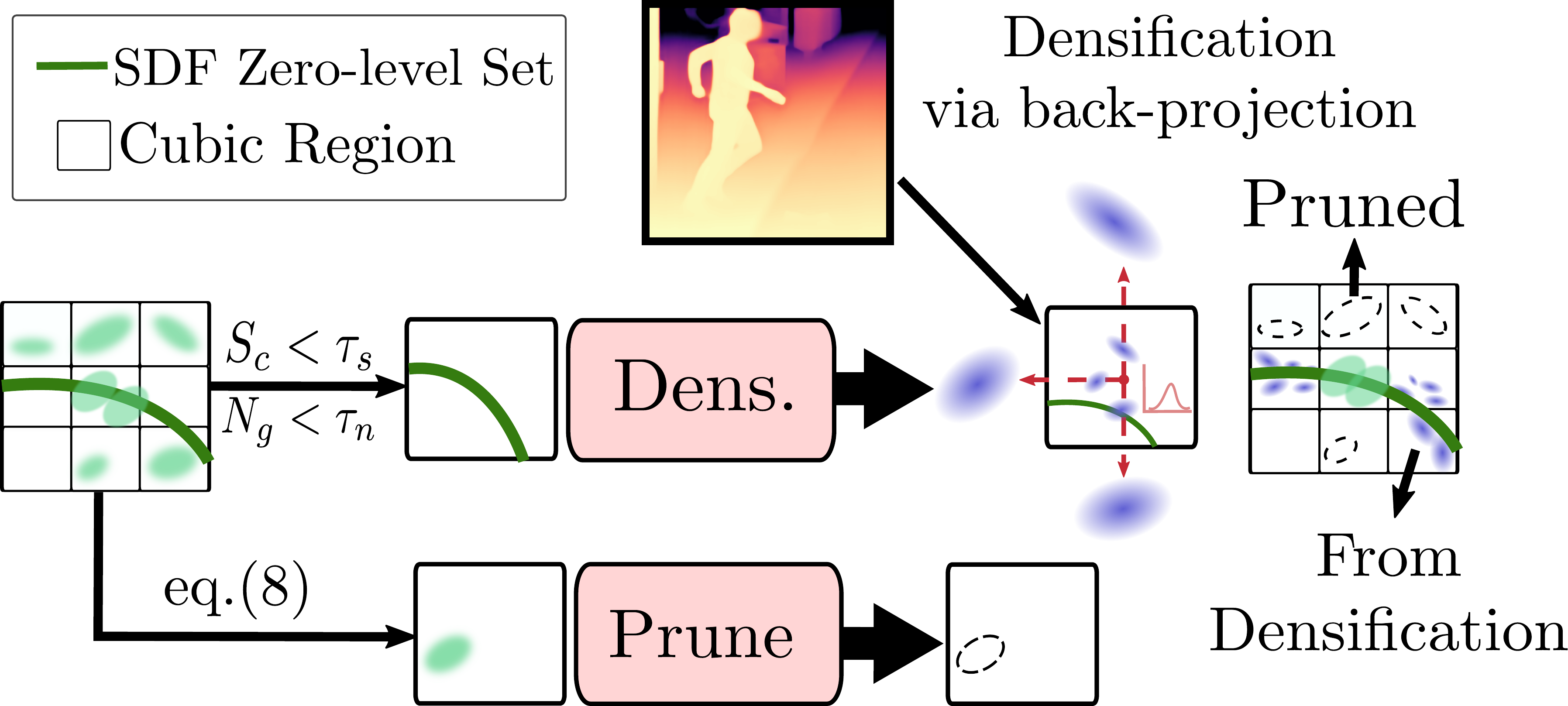}
    \caption{\textbf{SDF Guidance for Gaussian Primitive Distribution} Densification and Pruning of Gaussians is done by querying the SDF network. Points far away from the estimated SDF zero-level set are pruned, while sparse regions close to the zero-level set are chosen for densification. Dens. denotes densification.}
    \label{fig:edgs-sdf}
\end{figure}

Subsequently, we enumerate the Gaussians from $\mathcal{G}_t$ within each grid. In cases where the number of Gaussian primitives were insufficient $N_g < \tau_n$, we select the $K$ Gaussian neighbors of the grid's center point and generate $K$ new Gaussians within the grid. The initial attributes of these newly generated Gaussian primitives are sampled from a normal distribution defined by the mean and variance of the $K$ neighboring Gaussians.

\para{SDF Guided Pruning} For pruning, we integrate SDF information over a series of time-steps. If $\mathbf{x}_{i,t}$ is the position of a Gaussian at timestep $t$, we prune if 
\begin{equation}
    \sum_{t \in \text{prev timesteps}}\text{exp}(\frac{-S_i(t) + \sum_{j \in \text{NN}(i)}S_j(t)}{\gamma}) < \tau_{pr}
    \label{eq:prune}
\end{equation}
Here, $S_i(t)$ is the SDF value corresponding to $\mathbf{x}$ at frame $t$ and $\gamma$ is a hyper-parameter to prevent the exponent from reaching zero. Similarly,  $S_j(t)$ is the SDF value corresponding to a nearest neighbor of $\mathbf{x}_{i,t}$ at frame $t$ and $\gamma$ is a hyper-parameter to prevent the exponent from reaching zero. $t$ is taken over a small window of time. $j$ is taken over $K$ nearest neighbors. $\tau_{pr}$ is a threshold hyper-parameter to determine when to prune. \cite{xiang2024gaussianroom,yu2024gsdf} both propose pruning strategies that incorporate SDF and opacity information. In contrast, we don't use opacity as we found SDF values of a points neighbors to be a better indicator for pruning. %

\begin{table*}
\centering
\begin{tabular}{cccccccc}
\multirow{2}{*}{\textbf{Method}} &  \multirow{2}{*}{\textbf{Input}} & \multicolumn{3}{c}{\textbf{Scene Reconstruction}} & \multicolumn{3}{c}{\textbf{Novel View Synthesis}} \\
 &  & PSNR $\uparrow$ & SSIM $\uparrow$ & LPIPS $\downarrow$ & PSNR $\uparrow$ & SSIM $\uparrow$ & LPIPS $\downarrow$ \\
\hline

StreetGS~\cite{yan2024street} & \textit{M,T} & 29.11 & 0.921 & 0.117 & 25.71 & 0.764 & 0.218 \\
S3Gaussians~\cite{huang2024s3gaussians} & - & 31.35 & 0.911 & 0.106  & 26.82 & 0.788 & 0.226 \\
4DGF~\cite{fischer2024dynamic} & \textit{M,T} & 29.08 & 0.929 & 0.110 & 28.31 & \cellcolor{paleGreen}{0.859} & 0.206 \\
OmniRe~\cite{chen2024omnire} & \textit{T,M,S} & 33.79 & \cellcolor{paleGreen}{0.942} & \cellcolor{paleGreen}{0.105} & 29.35 & 0.780 & 0.186 \\
UGSDF (\textit{Ours}) & \textit{M,PT} & \cellcolor{paleBlue}{33.98} & \cellcolor{paleBlue}{0.944} & \cellcolor{paleBlue}{0.104} & \cellcolor{paleBlue}{30.63} & \cellcolor{paleBlue}{0.871} &  \cellcolor{paleBlue}{0.129} \\
UGSDF w/o LiDAR(\textit{Ours}) & \textit{M,PT} & \cellcolor{paleGreen}{33.88} & \cellcolor{paleGreen}{0.942} & \cellcolor{paleGreen}{0.105} & \cellcolor{paleGreen}{30.32} & \cellcolor{paleBlue}{0.871} &  \cellcolor{paleGreen}{0.145} \\
\end{tabular}%
\vspace{-0.3cm}
\caption{Evaluation on the Waymo Open Dataset. We evaluate each method in terms of PSNR, SSIM, and LPIPS for full-image quality. The \textbf{Input} column covers the different inputs used by the compared methods. \textit{T} denotes \textit{3D tracklet}, \textit{S} denotes \textit{SMPL}, \textit{M} denotes \textit{Masks}, \textit{PT} denotes point tracker. The \colorbox{paleBlue}{\textit{best}} and \colorbox{paleGreen}{\textit{second}} best results are highlighted. \textit{T}\& \textit{S} are both 3D priors while \textit{M} and \textit{PT} are 2D priors. }
\label{tab:waymo_comparison}
\end{table*}

\para{Gaussian-guided Point Sampling for SDF} Accurate surface reconstruction is achieved by sampling as close to the surface as possible. We employ the rasterized image from the 3D Gaussians to guide the point-sampling strategy. \cite{xiang2024gaussianroom,yu2024gsdf} propose using the rasterized depth maps of the Gaussians as a coarse geometry for point sampling. However, we found a better way was to input the rasterized image $\Hat{\mathbf{I}}_t$ into UniDepth and generate its aligned depth map $\Hat{\mathbf{D}}_t$ and use it  for guidance. Specifically, we leverage $\Hat{\mathbf{D}}_t$ from the 3D Gaussians to narrow down the ray sampling range.  Let $\mathbf{o}$ be the camera center and $\mathbf{d}$ be the viewing direction, we define the sampling range $\mathbf{r}$ of the SDF as:
\begin{equation}
    \mathbf{r} = [ \mathbf{o} + (\Hat{\mathbf{D}}_t(\mathbf{p}) - \gamma|S|).\mathbf{d}, \mathbf{o} + (\Hat{\mathbf{D}}_t(\mathbf{p}) + \gamma|S|).\mathbf{d} ]
\end{equation}
Here $\gamma$ is a hyper-parameter and $S$ is the  SDF value of the Gaussian that intersects with the ray.

\begin{table}\small
\centering
\resizebox{\columnwidth}{!}{ %
\begin{tabular}{@{}cS[table-format=2.2]S[table-format=1.3]S[table-format=1.3] cS[table-format=2.2]S[table-format=1.3]S[table-format=1.3]@{}}
\multirow{2}{*}{\textbf{Method}} & \multicolumn{3}{c}{\textbf{Scene Reconstruction}} & \multicolumn{3}{c}{\textbf{Novel View Synthesis}} \\
 & {PSNR $\uparrow$} & {SSIM $\uparrow$} & {LPIPS $\downarrow$} & {PSNR $\uparrow$} & {SSIM $\uparrow$} & {LPIPS $\downarrow$} \\
\toprule
StreetGS~\cite{yan2024street} & 26.73 & 0.883 & 0.162 & 25.61 & 0.803 & 0.211 \\
4DGF~\cite{fischer2024dynamic} & 27.16 & 0.885 & 0.149 & 26.47 & 0.761 & 0.237 \\
OmniRe~\cite{chen2024omnire} & \cellcolor{paleGreen}{27.95} & \cellcolor{paleGreen}{0.895} & \cellcolor{paleGreen}{0.147} & \cellcolor{paleGreen}{26.59} & \cellcolor{paleGreen}{0.816} & \cellcolor{paleGreen}{0.191} \\
UGSDF (\textit{Ours}) & \cellcolor{paleBlue}{29.55} & \cellcolor{paleBlue}{0.934} & \cellcolor{paleBlue}{0.105} & \cellcolor{paleBlue}{28.63} & \cellcolor{paleBlue}{0.926} & \cellcolor{paleBlue}{0.123} \\
\end{tabular}%
}
\caption{Performance comparison on the KITTI Dataset. We evaluate each method in terms of PSNR, SSIM, and LPIPS for full-image quality. The \colorbox{paleBlue}{\textit{best}} and \colorbox{paleGreen}{\textit{second}} best results are highlighted.}
\label{tab:kitti_comparison}
\end{table}

\begin{table*}\small
\centering
\begin{tabular}{lcccccccc}

\multirow{3}{*}{\textbf{Methods}} & \multicolumn{4}{c}{\textbf{Scene Reconstruction}} & \multicolumn{4}{c}{\textbf{Novel View Synthesis}} \\  
 & \multicolumn{2}{c}{Human} & \multicolumn{2}{c}{Vehicle} & \multicolumn{2}{c}{Human} & \multicolumn{2}{c}{Vehicle} \\ 
 & PSNR↑ & SSIM↑ & PSNR↑ & SSIM↑ & PSNR↑ & SSIM↑ & PSNR↑ & SSIM↑ \\
 \toprule
EmerNeRF\cite{yang2023emernerf} & 22.88 & 0.578 & 24.65 & 0.723 & 20.32 & 0.454 & 22.07 & 0.609 \\
DeformGS\cite{yang2024deformable} & 17.80 & 0.460 & 19.53 & 0.570 & 17.30 & 0.426 & 18.91 & 0.530 \\
PVG\cite{chen2023periodic} & 24.06 & 0.703 & 25.02 & 0.787 & 21.30 & 0.576 & 22.28 & 0.679 \\
StreetGS\cite{yan2024street} & 16.83 & 0.420 & 27.73 & 0.880 & 16.55 & 0.393 & 26.71 & 0.846 \\
OmniRe~\cite{chen2024omnire} & \cellcolor{paleBlue}{28.15} & \cellcolor{paleBlue}{0.845} & 28.91 & 0.892 & 24.36 & 0.727 & 27.57 & \cellcolor{paleGreen}{0.858} \\

UGSDF (Ours) & \cellcolor{paleGreen}{27.89} &\cellcolor{paleGreen}{0.839} & \cellcolor{paleBlue}{30.34} & \cellcolor{paleBlue}{0.906} & \cellcolor{paleBlue}{25.48} & \cellcolor{paleBlue}{0.758} & \cellcolor{paleBlue}{28.68} &  \cellcolor{paleBlue}{0.872}\\
UGSDF w/o LiDAR (Ours) & 27.89 & 0.825 & \cellcolor{paleGreen}{29.91} & \cellcolor{paleGreen}{0.902} & \cellcolor{paleGreen}{25.14} & \cellcolor{paleGreen}{0.742} & \cellcolor{paleGreen}{28.23} &  0.847\\
\end{tabular}
\vspace{-0.3cm}
\caption{\textbf{Comparison of methods for Scene Reconstruction and Novel View Synthesis on Human and Vehicle categories} The \colorbox{paleBlue}{\textit{best}} and \colorbox{paleGreen}{\textit{second}} best results are highlighted.}
\label{tab:human-vehicle}
\end{table*}

\begin{table}\small
\centering
\begin{tabular}{lcccc}
\multirow{2}{*}{\textbf{Methods}} & \multicolumn{2}{c}{\textbf{Human}} & \multicolumn{2}{c}{\textbf{Vehicle}} \\  
 & PSNR↑ & SSIM↑ & PSNR↑ & SSIM↑ \\
 \toprule
Ours (Full) & 27.89 & 0.839 & 30.34 &  0.916  \\
 w/o \textit{SG4GP}  & 22.47  &  0.541  & 22.27  & 0.612 \\
w/ \textit{Spars.} &  24.82 & 0.715  &  25.14 & 0.794 \\
w/o \textit{GRS4S} & 25.82 &  0.762 & 27.83  & 0.863 \\
\end{tabular}
\vspace{-0.3cm}
\caption{\textbf{Ablation Analysis}  on the NOTR dataset.}
\label{tab:ablation}
\end{table}
\begin{figure}
    \centering
    \includegraphics[width=0.97\linewidth]{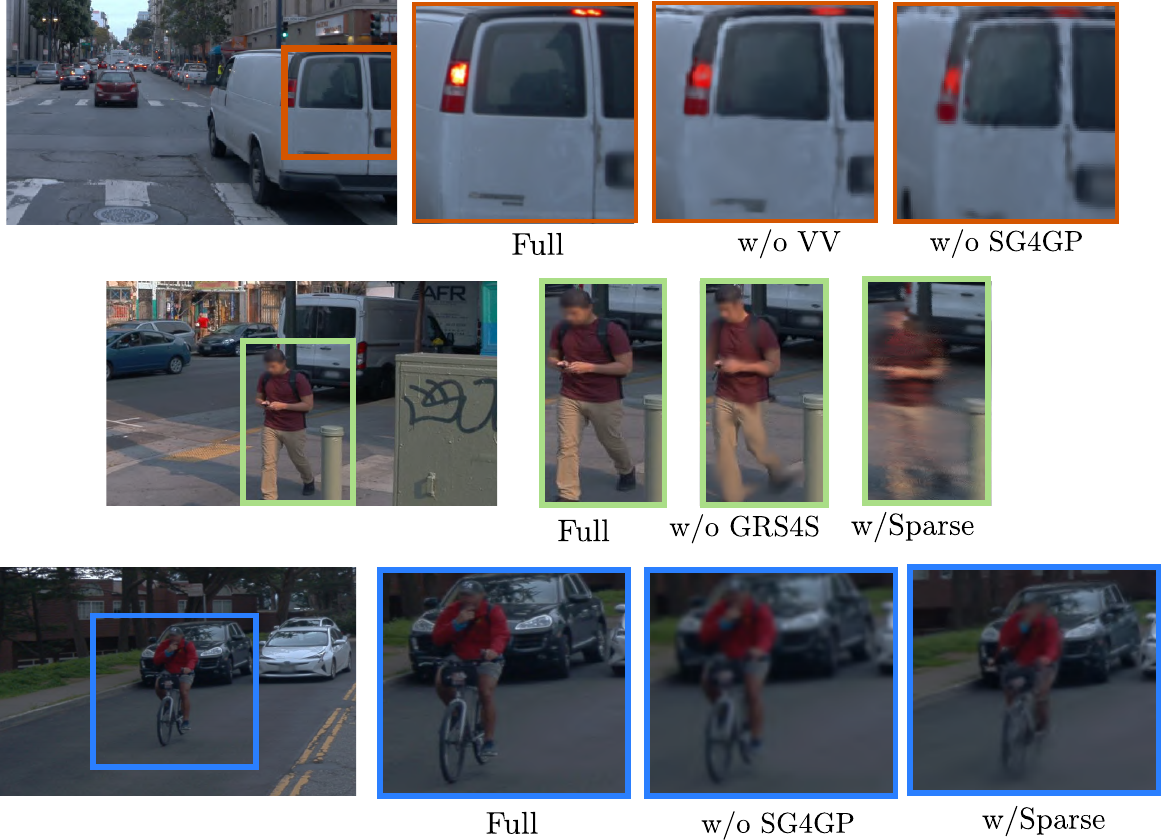}
    \vspace{-0.3cm}
    \caption{\textbf{Ablation Analysis} Our ablation analysis confirms the need for dense representations of thin objects like pedestrians and cyclists (rows 2 and 3). Furthermore, removing SDF guidance for Gaussian primitive distribution adversely impacts rendering quality (rows 1 and 3).}
    \label{fig:ablation}
\end{figure}
\section{Experiments}
\para{Baseline methods} We compare our method to four recent approaches for the tasks of full image reconstruction and novel-view synthesis: OmniRe~\cite{chen2024omnire}, 4DGF~\cite{fischer2024dynamic}, StreetGS~\cite{yan2024street}, and S3Gaussians~\cite{huang2024s3gaussians}. Additionally, for evaluating metrics on humans and vehicles, we compare our method against Periodic Vibration Gaussians (PVG)~\cite{chen2023periodic}, DeformGS~\cite{yang2024deformable}, and EmerNerf~\cite{yang2023emernerf}. 
Since our method specifically focuses on modeling dynamic objects, we handle the background and sky in a manner similar to OmniRe~\cite{chen2024omnire}; specifically, we \textit{minimize the same sky and background loss}. Additionally, we use SAM2~\cite{ravi2024sam} to generate masks for dynamic objects captured by the cameras, assigning a unique object ID to label the same object across different frames and camera views. Our method only uses tracks from~\cite{karaev2023cotracker} and depth from UniDepth~\cite{piccinelli2024unidepth}.

\begin{figure*}
    \centering
\includegraphics[width=0.97\linewidth]{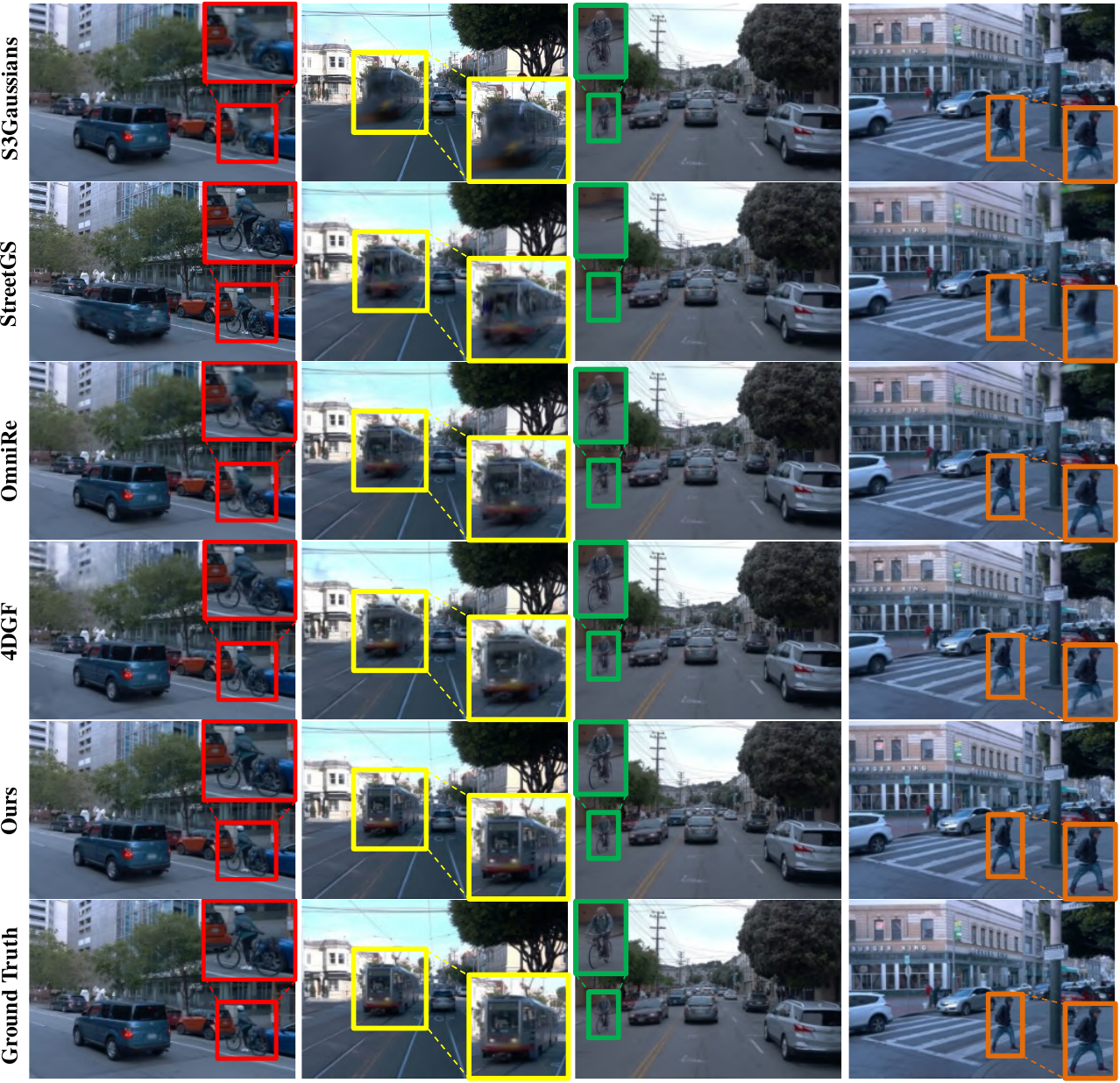}
    \vspace{-0.3cm}
    \caption{\textbf{Qualitative comparison of baselines and our method on the NOTR Dataset with zoomed in regions} We show results on some atypical objects observed in urban scenes namely: a tram and two cyclists, as well as a pedestrian.  Some of these are tricky cases not always very well modeled by some or all of the baselines. In contrast UGSDF achieves a very high level of fidelity in novel view synthesis for these objects.}
    \label{fig:qual}
\end{figure*}
\para{Datasets and Metrics} We evaluate our approach on two benchmarks: the Nerf-On-The Road (NOTR) 
 dataset~\cite{yang2023emernerf} (a subset of the Waymo Open Dataset~\cite{sun2020scalability}) and the KITTI MOT dataset~\cite{geiger2013vision}. For NOTR, we use the dynamic-32 split of ~\cite{yang2023emernerf}. It consists of  highly complex dynamic scenes that include
typical vehicles, pedestrians, cyclists and even a tram. For KITTI, we evaluate on the MOT 21 sequences of the city category as both StreetGS and OmniRe both report numbers on sequences with 3D tracklets. For all methods, we test on every $8^{th}$ frame and train on the rest for both datasets. We use the standard metrics of LPIPS, PSNR and SSIM to evaluate rendering quality. We show additional results on casually captured scenes from the iPhone dataset~\cite{gao2022dynamic} in the \textbf{supp}.

\subsection{Comparisons with State-of-the-art}
\cref{tab:waymo_comparison} and \cref{tab:kitti_comparison} shows the comparison of our approach on NOTR and KITTI respectively. Remarkably, our method outperforms methods like OmniRe and 4DGF by a small margin even though both use object pose information for vehicles in the form of bounding boxes. This holds true even if we don't make use of any LiDAR data in the pipeline for most metrics. This speaks to the robustness of 2D priors for view synthesis. OmniRe in-fact also uses SMPL templates and body pose predictors~\cite{goel2023humans}. Our method solely relies on depth from~\cite{piccinelli2024unidepth} and tracking information from~\cite{karaev2023cotracker}. \Cref{tab:human-vehicle} shows the results of our rendering on vehicles and humans. It shows that while OmniRe does do better on humans, it is not by a large margin. Secondly, our method is much better on vehicles. This explains the improvement our method has over OmniRe on the NOTR dataset as vehicles outnumber the pedestrians by in terms of pixels occupied. 

\Cref{fig:qual} shows qualitative results on the Waymo dataset. The results focus on non-rigid / atypical objects as all methods (baseline and ours) show reasonable results on most rigid objects. The first and third columns zoom in on  cyclists. OmniRe, S3Gaussians and StreetGS all struggle to model the cyclists, while the tram is poorly modeled by all methods except for UGSDF. While 4DGF models the cyclists well, it does a poor job on the background (check for column 1). For the fourth column, OmniRe and UGSDF are the only ones that capture all the details of the pedestrian. Whereas OmniRe, uses an SMPL mesh as a template, our method does so without the template.\textbf{We show additional image and depth results in the \textit{supp}}. We also show the \textbf{runtime analysis},  \textbf{tracking evaluation} of our method versus that of other methods in the \textbf{\textit{supp}, as well as \textit{scene-editing results}}.

\subsection{Ablation and Analysis}
\cref{tab:ablation} shows the results of the ablation analysis.
To understand which components contribute to the improvements in our method, we perform the following ablations: \textit{(1)} We disable \textit{SDF Guidance for Gaussian Primitives} (\textit{SG4GP}) and replace it with the standard adaptive density control used in the original 3DGS~\cite{kerbl20233d}. \textit{(2)} We disable \textit{Gaussian-guided Point Sampling for SDF} (\textit{GPS4S}). \textit{(3)} %
\textit{(3)} A default design choice in our method is to maintain a dense representation of dynamic objects based on SDF guidance. To evaluate this choice, we create a sparser representation by increasing the threshold $\tau_n$ for densification (denoted \textit{w/ sparse} in~\Cref{tab:ablation}). We show additional results and ablation in the \textbf{supp}.  

 Among the components, \textit{SG4GP} proves to be the most crucial, highlighting the importance of SDF representation for improving Gaussian primitive distribution. Maintaining a denser representation is also beneficial, especially for humans, showing a greater effect compared to vehicles. %
 
 \Cref{fig:ablation} presents qualitative results of the ablation study. As can be seen, removing SG4GP and using a sparser representation cause artifacting around the dynamic object especially thin ones like the  cyclist and pedestrian. The SDF network tends to learn relatively smooth surface representations. When combined with a sparse representation, this can result in under-fitting of the dynamic object. By using a denser representation the under-fitting is mitigated, enabling the SDF network to capture finer details of the object.

\section{Conclusions}
We propose a method combining Signed Distance Function (SDF) networks for precise geometry and 3D Gaussians for high-quality rendering. This joint learning approach improves the accuracy of each representation, achieving strong results on the NOTR and KITTI datasets by reconstructing both rigid and non-rigid dynamic objects and enabling novel view synthesis—all without ground truth motion, object templates, or 3D bounding boxes. Our method also supports tasks like object removal, scene composition, and novel view synthesis. Incorporating priors from video generative models would be a promising direction to explore. Also, we intend to explore jointly inferring motion segmentation~\cite{tourani2016using,tourani2024leveraging,tourani2024discrete}, in addition to scene rendering.

\paragraph{Limitations} UGSDF is sensitive to the 2D tracks generated by~\cite{karaev2023cotracker}, which can occasionally be inaccurate, leading to poor dynamic estimates of the Gaussians. Additionally, while the SDF network does provide some amount of controllability of the object, it does not have the representation power of the SMPL template. Finally, as with all the  baselines, UGSDF produces less
satisfactory novel views when the camera deviates significantly from the training trajectories.

{
    \small
    \bibliographystyle{ieeenat_fullname}
    \bibliography{main}
}

\maketitlesupplementary
\setcounter{section}{0}  %
\renewcommand{\thesection}{\Alph{section}}  %
\section{Losses for SDF Network}
\para{RGB and Depth Loss.}

The per-ray rendering losses for RGB and depth are defined as follows:

\begin{equation}
    \ell_{rgb}^r = \|\mathbf{c}_r^t - \mathbf{\hat{c}}_r^t\|, \quad \ell_d^r = |d_r^t - \hat{d}_r^t|
    \label{eq:rgbdlosses}
\end{equation}

where:
\begin{itemize}
    \item $\mathbf{c}_r^t$ is the observed RGB value of ray $r$ in the image.
    \item $d_r^t$ is the observed depth value of ray $r$ in the corresponding depth map.
    \item $\mathbf{\hat{c}}_r^t$ is the predicted RGB value for ray $r$.
    \item $\hat{d}_r^t$ is the predicted depth value for ray $r$ obtained from sampling.
\end{itemize}
\para{SDF supervision.} Following prior works \cite{Azinovic_2022_CVPR, isdf2022, wang2022go-surf}, we approximate the ground truth signed distance function (SDF) value using the distance to the observed depth along the ray direction $\mathbf{d}^t_r$. Specifically, we define the bound as 
\[
b_r(\mathbf{x}^t_i) = d^t_r - d(\mathbf{x}^t_i),
\]
where $d^t_r$ is the observed depth along ray $r$, and $d(\mathbf{x}^t_i)$ represents the depth of the sampled point $\mathbf{x}^t_i$.

Using this bound, we divide the sampled points into two disjoint sets:
\begin{itemize}
    \item \textbf{Near-surface points:} $S^{r}_{tr} = \{\mathbf{x}^t_i \mid b_r(\mathbf{x}^t_i) \leq \epsilon\}$, where $\epsilon$ is a truncation threshold that determines proximity to the surface.
    \item \textbf{Free-space points:} $S^{r}_{fs} = \{\mathbf{x}^t_i \mid b_r(\mathbf{x}^t_i) > \epsilon\}$, which are points far from the surface.
\end{itemize}

For the set of near-surface points $S^{r}_{tr}$, we define the following SDF loss to encourage accurate SDF predictions near the surface:
\begin{equation}
\mathcal{L}^r_{sdf} = \frac{1}{|S^{r}_{tr}|} \sum_{\mathbf{x}_s \in S^{r}_{tr}} \left| \varphi(\mathbf{x}^t_s) - b_r(\mathbf{x}^t_s) \right|,
\end{equation}
where $\varphi(\mathbf{x}^t_s)$ is the predicted SDF value at point $\mathbf{x}^t_s$.

For the set of free-space points $S^{r}_{fs}$, we apply a free-space loss similar to \cite{isdf2022, wang2022go-surf} to encourage free-space prediction and provide more direct supervision than the rendering terms in Eq.~\eqref{eq:rgbdlosses}:
\begin{equation}
\mathcal{L}_{fs}^{r} = \frac{1}{|S_{fs}^{r}|} \sum_{\mathbf{x}_s \in S_{fs}} 
\max \left( 
    0, 
    e^{-\alpha \varphi(\mathbf{x}_{s}^{t})} - 1, 
    \varphi(\mathbf{x}_{s}^{t}) - b_{r}(\mathbf{x}_{s}^{t})
\right).
\end{equation}

This loss applies:
\begin{itemize}
    \item An exponential penalty for negative SDF values.
    \item A linear penalty for positive SDF values exceeding the bound.
    \item No penalty when the SDF value is within the bound.
\end{itemize}

\para{SDF regularization.} To ensure valid SDF values, particularly in regions without direct supervision, we incorporate the Eikonal regularization term $\ell_{eik}$, which promotes a uniform gradient norm for the SDF, encouraging it to grow smoothly away from the surface \cite{pmlr-v119-gropp20a, isdf2022, NEURIPS2021_neus}. Specifically, for any query point $\mathbf{x}_i^{\prime t}$ in the canonical space $\mathbb{R}^3$, the gradient of the SDF with respect to the 3D point is encouraged to have unit length:

\begin{equation}
\mathcal{L}_{eik}^{r} = \frac{1}{|S_{fs}^{r}|} \sum_{\mathbf{x}_s \in S_{fs}} \left(1 - \|\nabla \varphi(\mathbf{x}_s^{\prime t})\|\right)^2.
\end{equation}

\para{Surface smoothness regularization.} To enhance surface smoothness, we enforce nearby points to have similar normals. Unlike \cite{wang2022go-surf}, which samples uniformly within a grid, we sample only surface points $\mathbf{x}_s^t \in S_{surf}$, significantly reducing computation. The smoothness loss is defined as:

\begin{equation}
    \mathcal{L}_{sm} = \frac{1}{R} \sum_{\mathbf{x}_s \in S_{surf}} \|\nabla \varphi(\mathbf{x}_s^t) - \nabla \varphi(\mathbf{x}_s^t + \delta)\|^2,
    \label{eq:smoothness}
\end{equation}

where $\mathbf{x}_s^t$ is back-projected using depth maps, $\delta$ is a small perturbation sampled from a Gaussian distribution with standard deviation $\delta_{std}$, and $R$ is the total number of sampled rays.

The RGB rendering loss $\mathcal{L}_{rgb}$ measures the difference between ground truth and predicted ray colors, while the depth rendering loss $\mathcal{L}_d$ evaluates the depth error over valid rays $R_d$. Both losses utilize the object mask $M_r$ to focus on the object of interest:

\begin{equation}
    \mathcal{L}_{rgb} = \frac{1}{|R_{rgb}|} \sum_{r \in R_{rgb}} M_r^t \ell_{rgb}^r,
    \label{eq:rgb_loss}
\end{equation}

\begin{equation}
    \mathcal{L}_{d} = \frac{1}{|R_{d}|} \sum_{r \in R_{d}} M_r^t \ell_{d}^r.
    \label{eq:depth_loss}
\end{equation}

The SDF loss $\mathcal{L}_{sdf}$ is applied to points in the truncation region $S_{tr}$:

\begin{equation}
    \mathcal{L}_{sdf} = \frac{1}{|R_d|} \sum_{r=1}^R \mathcal{L}_{sdf}^r.
    \label{eq:sdf_loss}
\end{equation}

The free-space loss $\mathcal{L}_{fs}$ and Eikonal loss $\mathcal{L}_{eik}$ are applied to the remaining points $S_{fs}$:

\begin{equation}
    \mathcal{L}_{fs} = \frac{1}{|R_d|} \sum_{r=1}^R \mathcal{L}_{fs}^r,
    \label{eq:fs_loss}
\end{equation}

\begin{equation}
    \mathcal{L}_{eik} = \frac{1}{|R_d|} \sum_{r=1}^R \mathcal{L}_{eik}^r.
    \label{eq:eik_loss}
\end{equation}

\section{Dataset Details}
We evaluate our method on the NOTR Dataset~\cite{yangemernerf}, which uses sequences from the Waymo Open Dataset~\cite{sun2020scalability}. The scene IDs used in our experiments are listed in Table~\ref{tab:dynamic-scenes}.

\begin{table}[t]
\centering
\begin{tabular}{cccccccc}
\toprule
\toprule
16 & 21 & 22 & 25 & 31 & 34 & 35 & 49\\
53 & 80 & 84 & 86 & 89 & 94 & 96 & 102\\
111 & 222 & 323 & 382 &402 & 427 & 438 & 546\\
581 & 592 & 620 & 640 & 700 & 754 & 795 & 796\\
\bottomrule
\bottomrule
\end{tabular}
\caption{Scene IDs of 32 dynamic scenes from the  NOTR~\cite{yangemernerf} Dataset which is a subset of the Waymo dataset used for evaluation.}
\label{tab:dynamic-scenes}
\end{table}

We also evaluate on the KITTI MOT sequences for which 3D tracklets are available, as the other two baselines~(\cite{chen2024omnire,fischer2024dynamic}) utilize these tracklets. The specific sequence IDs used for this evaluation can be seen in Table~\ref{tab:kitti-sequences}
\begin{table}[t]
\centering
\scriptsize
\begin{tabular}{cc}
\hline
2011\_09\_26\_drive\_0005 (City) & 2011\_09\_26\_drive\_0009 (City) \\
2011\_09\_26\_drive\_0011 (City) & 2011\_09\_26\_drive\_0013 (City) \\
 2011\_09\_26\_drive\_0014 (City) & 2011\_09\_26\_drive\_0015 (Road)\\
  2011\_09\_26\_drive\_0018 (City) & 2011\_09\_26\_drive\_0022 (Residential) \\
2011\_09\_26\_drive\_0032 (Road) & 2011\_09\_26\_drive\_0036 (Residential) \\  2011\_09\_26\_drive\_0056 (City) & 2011\_09\_26\_drive\_0059 (City) \\  2011\_09\_26\_drive\_0060 (City) & 2011\_09\_26\_drive\_0091 (City)  \\
\hline
\end{tabular}
\caption{KITTI raw sequences.}
\label{tab:kitti-sequences}
\end{table}

\section{Runtime-Analysis}
We show the runtime analysis for our method relative to other methods in~\cref{tab:runtime-analysis}. Our method is competitive with other methods both in terms of frame rate and training time. This is because the background Gaussians take up the most amount of time for the rendering operation. 

Training for our method takes 3-5 hours for a single sequence. $60\%$ of the time is typically taken to 
train the SDF networks, $5\%$ for initialization, with the remaining $35\%$ by rasterization approximately. At inference 
time, our method runs at about 20 fps, which is similar to
4DGF~\cite{fischer2024dynamic} and S3Gaussians~\cite{huang2024s3gaussians}.

\addtolength{\tabcolsep}{-0.4em}
\begin{table}[h]
\scriptsize
    \centering
    \begin{tabular}{ccccccc}
         \toprule
         Method & Ours & OmniRe & S3Gaussians & 4DGF & StreetGS\\
         & & \cite{chen2024omnire} & \cite{huang2024s3gaussians} & \cite{fischer2024multi} & \cite{yan2024street}\\ 
         \midrule 
     Train Time    & 3-5  & 3-5 & 8-10 & 3-5 & 1-2\\
        Frame Rate  & 20 & 24 & 20 & 20 & 68\\
         \bottomrule
    \end{tabular}
    \caption{Train time (in hrs) and frame rate (in fps) comparison for our method. }
    \label{tab:runtime-analysis}
\end{table}

\section{Tracking Evaluation}
To evaluate the tracking off the combination of the depth network UniDepth~\cite{piccinelli2024unidepth} and the point tracker CoTracker V3~\cite{karaev2023cotracker}, we performing a tracking evaluation. To do so, we compare the rendering results of using bounding boxes derived from our tracking methodology, to that of~\cite{pang2022simpletrack} and the ground truth on the KITTI MOT dataset. Remarkably the combination of point tracking and depth yields only slightly inferior results to that of GT bounding boxes while being almost identical to~\cite{pang2022simpletrack} which makes an assumption of object rigidity.

\begin{table}[]
\small
    \centering
    \begin{tabular}{cccc}
       3D BBox Type  & PSNR & SSIM & LPIPS  \\
       \midrule
       GT  & 31.34 & 0.945 & 0.026\\
       Ours & 30.55 & 0.931 & 0.028\\
       \cite{pang2022simpletrack} & 30.67 & 0.943 & 0.035\\
    \end{tabular}
    \caption{Comparison of GT Bounding Boxes (GT), bounding boxes predicted from our tracking method (Ours) and from~\cite{pang2022simpletrack}.}
    \label{tab:my_label}
\end{table}

\section{Additional Results}
\subsection{Image and Depth Rendering  Results} 
\begin{figure*}
\centering
    \includegraphics[width=0.95\linewidth]{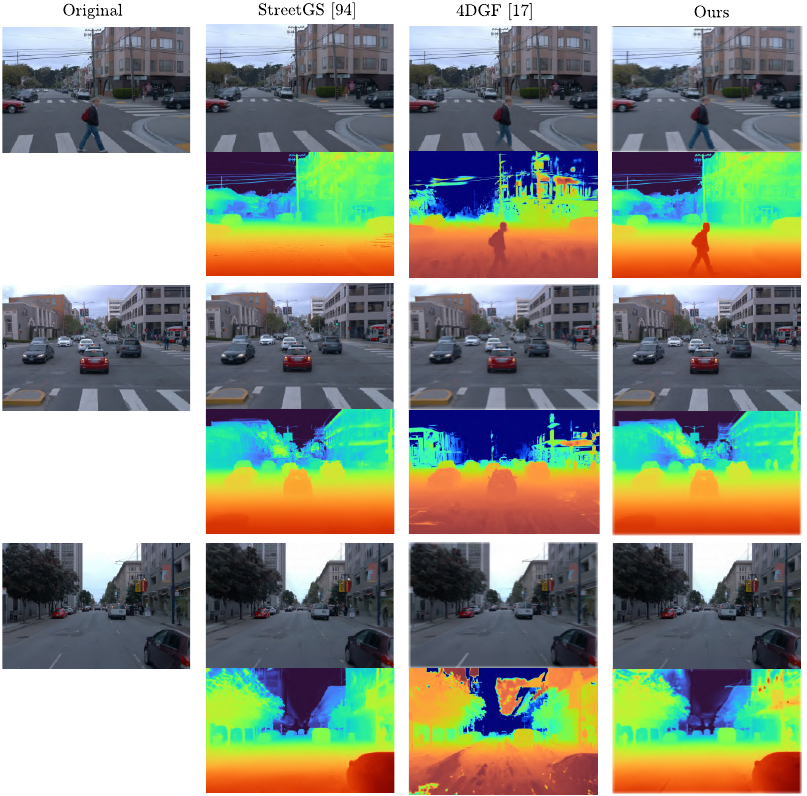}
\caption{\textbf{Image and Depth Rendering Results for the NOTR Dataset} Our method is rendered without LiDAR and is compared to StreetGS and 4DGF. Even though the rendered images look similar, the depth achieved varies by mehod. Our method is able to capture the details (feet of the pedestrian)  and smoothness of moving objects (cars) with greater accuracy. StreetGS cannot model pedestrians, hence it fails to render in the top strip. \textit{Citation numbers in the figure correspond to the main paper.}}
\label{fig:render}
\end{figure*}

\Cref{fig:render} shows the results of rendered scenes for both depth and images with moving objects for StreetGaussians, 4DGF and our method without LiDAR inputs. Using point tracking and depth only, our network is able to preserve the details of moving objects in greater detail than that of methods that make use of LiDAR.

\subsection{Scene Editing Results} 
\begin{figure*}[h!]
\centering
\includegraphics[width=0.80\linewidth]{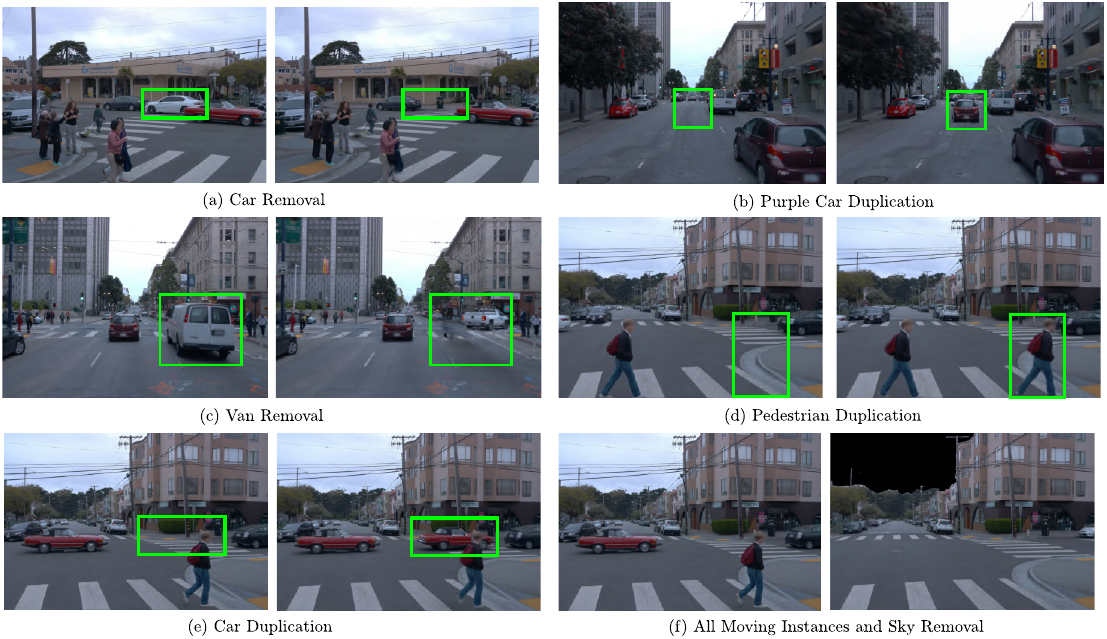}
    \caption{\textbf{Scene Editing} We show original and edited pairs of images, with the region of interest highlighted in a green bounding box. (a) and (c) show the white car and van removed from the scene respectively. (b), (d) and (e) show duplicated cars and pedestrians in the scenes. (f) shows the rendered scene after removing all moving objects and the sky. Videos for the edited scenes are in the \textbf{\textit{supp. video}}.}
    \label{fig:scene-edit}
\end{figure*}

\Cref{fig:scene-edit} shows some scene editing results on different video sequences. We are able to add and remove both rigid and non-rigid objects from the scene. (b), (d) and (e) add a car, pedestrian and car respectively. (a), (c) and (f) remove instances.

\subsection{Results on IPhone Dataset} As our method is not restricted to urban scene datasets but can work on more casually captured datasets, we show qualitative results on the IPhone Dataset. We compare the results to Shape Of Motion~\cite{som2024}, Dynamic Gaussian Marbles~\cite{stearns2024marbles}. 

\begin{figure*}[t]
\centering
    \includegraphics[scale=0.80]{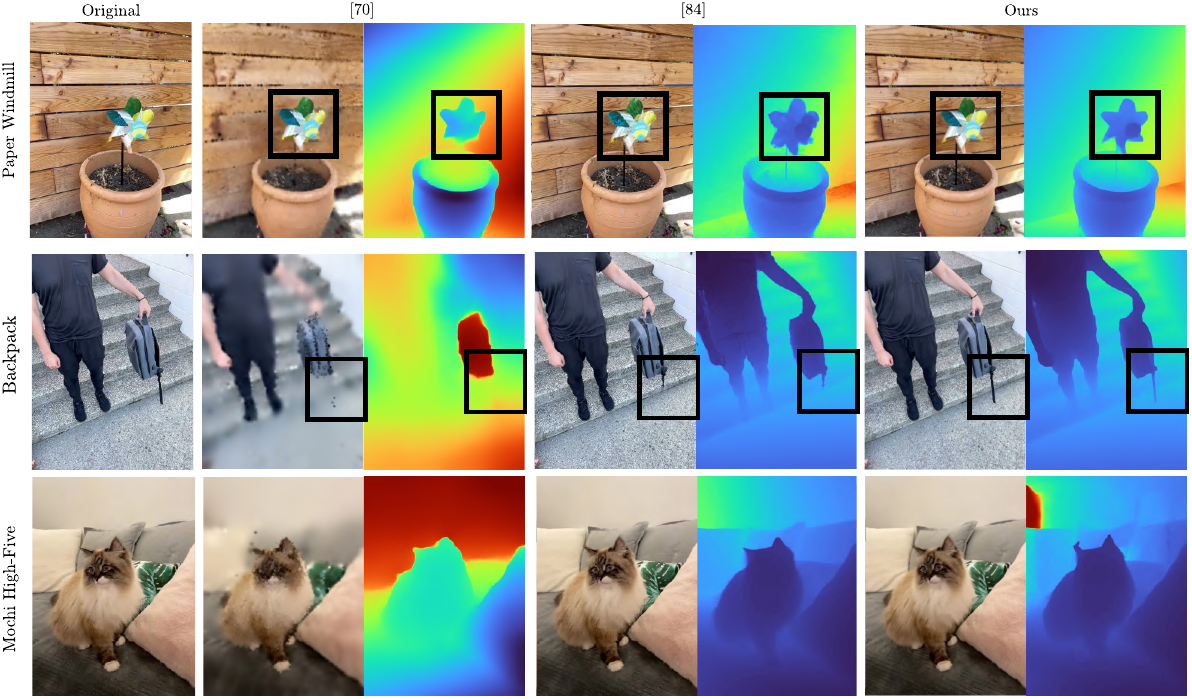}
    \caption{\textbf{Qualitative Results on the IPhone Dataset} We show the rendering results of Dynamic Gaussian Marbles~\cite{stearns2024marbles}, Shape of Motion~\cite{som2024} and our method on 3 sequences of the IPhone Dataset. The bounding boxes highlight regions where our method generates a more high-fidelity rendering of the scene. \textit{Citation numbers in the figure correspond to the main paper.}}
\end{figure*}

\section{Implementation Details}
\para{Initialization} For the background model, we follow OmniRe~\cite{chen2024omnire}, combining LiDAR points
with $4 \times 10^5$ random samples, which are divided into $2 \times 10^5$ near samples uniformly distributed by distance to the scene’s origin and $2 \times 10^5$ far samples uniformly distributed by inverse distance. To initialize the background, we filter out the LiDAR samples of dynamic objects. For canonicalization around dynamic objects, we use the depth map estimated from UniDepth to  calculate a bounding box around the object. We use the 2D tracks generated from~\cite{karaev2023cotracker} to warp lidar and depth information from neighboring frames into the initialization frame, typically chosen as the frame where the object is initially detected via SAM2~\cite{ravi2024sam}.

\para{Optimization} Our 3DGS pipeline trains for $30,000$ iterations with all scene nodes optimized jointly. The learning
rate for Gaussian properties aligns with the default settings of 3DGS~\cite{kerbl3Dgaussians}. Instead of using spherical harmonics, we just use a constant color value for the Gaussians. 
For the SDF Network, for the initialization, we train the network for 2000 iterations. We train at the lowest-resolution for the first 500 iterations, adding an additional resolution every 200 iterations during the initialization. Subsequently, we train the iteration a 1000 iterations every 2000 training iterations for the Gaussians. We train the SDF network using the Adam optimizer~\cite{kingma2014adam} with a learning rate $5\times 10^{-4}$. As mentioned in the main paper training alternates between the SDF network and the Gaussians and is done progressively. We employ step-based weighting for the RGB, depth, and regularization losses, prioritizing RGB and regularization losses early in training and gradually reducing their weights as training progresses. To begin with, we randomly sample an image and select 1024 rays per batch, sampling 128 points along each ray. Subsequently, we take feedback from the Gaussian splatting to direct the ray sampling improving upon random ray sampling. Overall optimization time for our is around 3-4 hours per scene. 

\para{Hyper-Parameters}:

\begin{itemize}
    \item \textbf{SDF Guidance for Gaussians} $\tau_s$ = 0.01, $\tau_n$ = 0.02, $\tau_{pr}$=0.02
    \item \textbf{Gaussian Guidance for SDFs}: $\gamma=3$
\end{itemize}

\end{document}